\documentclass[10pt,twocolumn,letterpaper]{article}

\usepackage{cvpr}
\usepackage{times}
\usepackage{epsfig}
\usepackage{graphicx}
\usepackage{amsmath}
\usepackage{amssymb}
\usepackage{amsthm}
\usepackage{bm}
\usepackage{placeins}
\usepackage[ruled,vlined]{algorithm2e}

\cvprfinalcopy 
\ifcvprfinal\fi

\newtheorem{defn}{Definition}[section]

\def\cvprPaperID{1730} 

\def\q{\mathbf{q}}

\begin{document}

\title{In the Shadows, Shape Priors Shine:\\ Using Occlusion to Improve Multi-Region Segmentation}

\author{Yuka Kihara\\
Ricoh Co., Ltd.\thanks{Work done while the first author was visiting Cornell University.}\\
{\tt\small yuuka.kihara@nts.ricoh.co.jp}
\and
Matvey Soloviev\\
Cornell University\\
{\tt\small msoloviev@cs.cornell.edu}
\and
Tsuhan Chen\\
Cornell University\\
{\tt\small tsuhan@cornell.edu}
}

\maketitle

\begin{abstract}
We present a new algorithm for multi-region segmentation of 2D images with objects that may partially occlude each other. Our algorithm is based on the observation that human performance on this task is based both on prior knowledge about plausible shapes and taking into account the presence of occluding objects whose shape is already known -- once an occluded region is identified, the shape prior can be used to guess the shape of the missing part. We capture the former aspect using a deep learning model of shape; for the latter, we simultaneously minimize the energy of all regions and consider only unoccluded pixels for data agreement. 

Existing algorithms incorporating object shape priors consider every object separately in turn and can't distinguish genuine deviation from the expected shape from parts missing due to occlusion. We show that our method significantly improves on the performance of a representative algorithm, as evaluated on both preprocessed natural and synthetic images. Furthermore, on the synthetic images, we recover the ground truth segmentation with good accuracy.
 
\end{abstract}

\section{Introduction}

The objective of multi-region segmentation of an image is to determine the boundaries of a number of regions containing objects of interest. Unlike the single-region case, in which the boundary unambiguously defines a partition of the image domain (and so every pixel can be associated with the object or the background), region membership becomes ambiguous when multiple regions are considered and are allowed to intersect. While several methods for multi-region segmentation have been presented in the past \cite{c26,c27,c28,c29, c39}, most of these algorithms generate a true partition of the image into regions (including the background), so that every image point is assigned to a unique region.

 
The scenario where regions may overlap arises naturally when trying to infer the true outlines of a number of objects in a two-dimensional picture of a 3D scene, which may partially occlude each other. Human observers have been known to vastly outperform known algorithms in this setting \cite{c30, c31}; folk wisdom and previous studies \cite{c1,c2,c3,c4,c5,c6} partially attribute this gap to extensive explicit (having seen many objects, familiarity with laws of physics) and implicit (neurovisual tendency towards completing straight lines, constant-curvature curves and many more complex features) prior knowledge about plausible object shapes. Yet, as the following pair of figures demonstrates, this is not the whole story: 

\begin{figure}[h]
\begin{center}
\fbox{\includegraphics[width=0.25\linewidth]{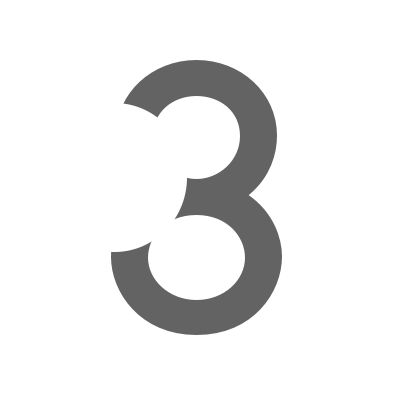}}
\fbox{\includegraphics[width=0.25\linewidth]{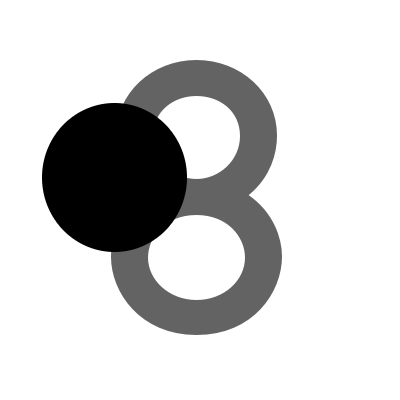}}
\end{center}
\vspace{-12pt}
\label{fig:threevseight}
\end{figure}
On the right, we clearly see a figure ``8'', partially obscured by a black circle in front of it. The figure on the left, though, could be resolved either in the same fashion (especially if the background is not a uniform color), or as an unusually-shaped figure ``3'', with most observers, human or algorithmic, likely to lean towards the latter.

This suggests that besides prior knowledge about the shape of objects, knowledge of the presence and shape of occluding objects is also relevant to segmentation performance -- failure of the image to manifest the expected shape of a figure ``8'' due to the presence of an overlapping shape counts as weaker evidence against the presence of a figure ``8'' than failure due to a gap or perhaps noise. In fact, it is precisely in the regions that can be established to be occluded that prior knowledge about shape attains its full significance -- in the absence of data on occluded pixels, whether a certain class of object may be identified in a region depends on whether the exposed parts may be extended into an occluded region in a new way that creates a shape considered plausible for that class.
 
In this paper, we will explore to what extent taking both presence of other objects and prior knowledge about shapes into account may allow us to improve upon existing multi-region segmentation algorithms. To this end, we shall introduce an algorithm for multi-region segmentation that optimizes both for adherence to a machine-learned prior notion of plausible shapes (the shape prior) and agreement with the image data where it is appropriate -- that is, where the object presumed to lie in the region is not occluded by some other object situated in front of it. In particular, this does entail simultaneously evolving multiple partially overlapping region boundaries. To our knowledge, no previous attempts \cite{c1,c2,c3,c4,c5,c6,c13,c14,c15,c21,c22} at multi-region segmentation attempted to employ shape priors in this fashion.

Our ``plausible shapes'' will be drawn from one of a number of classes of familiar objects, each represented using the Shape Boltzmann Machine deep learning model (introduced in Section \ref{sec:shapebm}). We interpret the problem as an extension of Nitzberg, Mumford and Shiota's \emph{Segmentation with Depth} \cite{c7, c8} (Section \ref{sec:nms}), formulated with a data-driven term that favors uniformly colored regions of the input region and a shape term calculated by the machine learning model from a spectral descriptor representation of the region; we introduce a new way to combine and optimize for both of them simultaneously using the NMS functional in Section \ref{sec:probrep}.
 
\section{Related Work}
As previously mentioned, the segmentation with depth problem was previously formulated by Nitzberg, Mumford and Shiota in \cite{c7, c8} using a variational formulation.
The NMS model consists of a sum of a data driven term of each region, which favours segmenting curves close to the boundaries of visible parts, and a shape constraint term that constrains the possible shape contour of objects inside the image, including occluded parts.
Segmentation with depth thus determines the boundaries of overlapping objects, taking into account the shape and the relative distance(depth) order of the objects, based on intensity distributions in the object regions.
Numerical methods for minimizing the NMS model have been presented in \cite{c9, c10, c11}. 
In \cite{c12}, multiple shape prior segmentation task using graph cuts has been presented. This method is capable of handling overlap by allowing for a pixel to have multiple object memberships (labels) and simultaneously segmenting multiple objects. They define the shape prior energy using a discrete version of the shape distance proposed in \cite{c11} for the level sets framework, and incorporate this energy into the graph via terminal edge weights. 

However, all these works consider very limited shape priors such as contour smoothness or directly known objects, each class consist of only one image. It is not adequate to represent the shape prior encoding more complex shape variation. 
Recently, deep learning models \cite{c13,c14} are attractive for their well performance in modeling high-dimensional richly structured data. 

A deep learning model is a machine learning model that considers multiple levels of representation and abstraction that help to make sense of data such as images, sound, and text. 
One such model, called a Shape Boltzmann Machine (SBM) \cite{c14}, is a particular form of Deep Boltzmann Machine (DBM)  \cite{c13} proposed specifically for the task of modeling binary object shapes. An SBM is trained in a generative manner: given a set of shapes, the goal is to learn a probabilistic model that models the given shapes accurately and can generalize to unseen instances of the multiple shape categories. Due to its structure, SBM is successfully reduce the number of first layer parameters and facilitate efficient learning for smaller datasets while preserving its ability.  

We use the SBM to incorporate information about shape in the NMS formulation. The shapes can be a set of different familiar objects, but train without information about image class. The learned model parameters implicitly define a probability distribution over all possible binary shapes in SBM; combining this prior with a variational segmentation model will necessitate a probabilistic rather than level-set representation of shape, which we will introduce in \ref{sec:probrep}.

\section{Overview: NMS Segmentation with Depth}
\label{sec:nms}
We shall briefly review the formulation of segmentation with depth by Nitzberg, Mumford and Shiota\cite{c7}.

Let ${\bf u}:\Omega \to \mathbb{R}^1$ be a two-dimensional greyscale image defined on image domain $\Omega \subset \mathbb{R}^2$.

We make the following assumptions about the scene:
\begin{enumerate}
\item No object to be recognised is completely occluded.
\item Every object is of approximately uniform brightness, and any two objects in the scene have different brightness.
\item The objects are not interlaced, so in particular there is a well-defined ordering by distance from the observer.
\end{enumerate}
Note that by ignoring invisible objects, all three assumptions are easily satisfied by considering depth images of spatially well-separated objects.

Let $R_1$, $\ldots$ , $R_n$ be the regions occupied by objects, ordered by increasing distance from the observer (i.e. depth). Each region $R_i$ may be partially occluded by the regions preceding it in the ordering (but not completely so, by assumption 1). We shall denote the visible parts of the above objects as $R'_1$, $\ldots$, $R'_n$ respectively; they may be formally defined as 
\begin{eqnarray*}
 R'_1 &=& R_1 \\
 R'_i &=& R_i - \bigcup_{j<i} R_j \text{ for $i>1$}.
\end{eqnarray*}
We may denote the background as $R'_{n+1} = \Omega - \bigcup_{j\leq n}R_j$ for the sake of consistency.

\begin{defn} (NMS) Given an image $f:\Omega\rightarrow \mathbb{R}$ and a number $n$ of recognisable objects, the objective of \emph{segmentation with depth} is to determine
\begin{itemize}
\item an ordering $O_1<\ldots<O_n$ of recognisable objects in the image by distance from the viewer;
\item the average greyscale intensity $c_i \in \mathbb{R}$ of the region occupied by each object $O_i$ and
\item the shape $R_i \subseteq \Omega$ of the region occupied by each object $O_i$. 
\end{itemize}
\end{defn}

The \emph{NMS functional} provides an objective function for comparing candidate solutions for the third point, given solutions for the first two. In its general form, it may be written as follows:
\begin{eqnarray}
E_\mathrm{NMS} &=& \sum _{i=1}^{n+1} \int _{R'_i} |{\bf u}(x,y)-c_i|^2 dxdy \nonumber \\ &+& E_\mathrm{shape}. \label{eq:nms_energy}
\end{eqnarray}


The $E_\mathrm{shape}$ term of $E_{NMS}$ serves to penalise implausible shapes of regions. In \cite{c7}, the authors set $$E_\mathrm{shape}=\sum_{i=1} ^{N} \oint_{\partial R_i \cap \Omega} [\alpha+\beta \psi(\kappa)] ds.$$ As this contour integral is taken along the boundary of the original unoccluded shape, evaluating the function inevitably entails reconstructing invisible parts of regions' boundaries. Depending on the setting of the parameters $\alpha$ and $\beta$, one may obtain curvature functions such as Euler's Elastica.
 
In our work, instead of a curvature-based boundary energy function, we use the evaluation of a deep learning model trained on global and local features of ``good''/``known'' shapes, which we shall briefly introduce in the next section, plus a denoising term based on the total variance norm as in \cite{c15}. The top-level segmentation algorithm then takes the following form:
\begin{enumerate}
\item By running $k$-means on the histogram of the input depth image ${\bf u}$, determine the number and average intensity of objects present. As intensity represents depth in the image, this also gives a canonical depth ordering.
\item For each object $i$, find a small uniform region of the average colour determined for it initialise $R_i$ with it.
\item Perform gradient descent on a relaxed version of the NMS functional using the Split Bregman method (section \ref{sec:decomposed}) to find a local minimum for $\{R_i\}_{1\leq i\leq n}$.
\end{enumerate}


\section{Algorithm}
\subsection{The Shape Boltzmann Machine}
\label{sec:shapebm}
In our work, we implement the shape energy term using a \emph{Shape Boltzmann machine} (SBM). (For an in-depth introduction of SBMs and DBMs, we refer the reader to \cite{c14} and \cite{c13} respectively.) Our SBM employs two layers of hidden variables. We shall denote by $\bf v$ the vector of visible units that represents the (input) binary shape image, and by $\bf h^1$ and $\bf h^2$ be the lower and higher binary hidden units respectively.  


The first layer $\bf h^1$ consists of several disjoint subsets $h_1^k$ of same size $m$. Each of these hidden units has a restricted receptive field and only connects to a subset of the visible units, but sharing weights between the sets of hidden units and visible units $\bf W^1 =W^1_k$, $k \in K$. The visible units are local square patches having the same size $n$ and overlapping its neighbor by $d$ pixels along the boundaries. 

 
Similar to DBMs, there are no within-layer connections. The (output) energy of the state $\bf \{v, h^1, h^2\}$ is then defined as follows:
\begin{eqnarray}
& &E_{SBM}({\bf v,h^1,h^2}; \theta) \nonumber\\
&=& -\sum_{k \in{K}} \bf v_k^{\mathrm{T}}W^1h_k^1 \nonumber\\
& & \bf - b^{\mathrm{T}}v -c^{1{\mathrm{T}}}h^1-h^{1\mathrm{T}}W^2h^2-c^{2\mathrm{T}}h^2
 \label{eq:energy_sbm}
\end{eqnarray} 

Here, $\theta = \{\bf \tilde{W}^1, W^2, c^1, c^2, b\}$ are the model parameters. $ \tilde{W}^1$ is defined as follows.
The first term can be rewritten in the same form of DBM by some matrix manipulation: 
\begin{equation}
 \sum_{ k \in{K}} \bf v_k^{\mathrm{T}}W^1h_k^1=v^{\mathrm{T}} \tilde{W}^1 h^1  
\end{equation}

The distribution over $v$ is given by marginalizing over the hidden variables.

\begin{eqnarray}
p({\bf v};\theta)=\frac{1}{Z(\theta)}\sum_{\bf h^1,h^2}exp(-E(\bf v,h^1, h^2)) 
 \label{eq:distribution_sbm}
\end{eqnarray} 

Here, $Z(\theta)$ is the partition function. Although exact inference is no longer tractable, an efficient approximate learning can be carried out by using a mean-field procedure based on the property that conditional distributions $p({\bf v} | {\bf h^1})$, $p({\bf h^1} | {\bf v, h^2})$, and $p({\bf h^2} | {\bf h^1})$ remain independent, as detailed in \cite{c13}. 
Each conditional distribution is given by sigmoid function $\sigma(y)=1/(1+exp(-y))$ as follows.

\begin{flalign}
p({h^1_j=1} | {\bf v, h^2})&=\sigma(\sum_i \tilde{W}^1_{ij}v_i + \sum_k W^2_{jk}h^2_k+c^1_j)\\
p({h^2_k=1} | {\bf h^1})&=\sigma (\sum_j W^2_{jk}h^1_j+c^2_k)\\
p({v_i=1} | {\bf h^1})&=\sigma(\sum_j \tilde{W}^1_{ij}h^1_j+b_i) 
 \label{eq:conditional_distribution_sbm}
\end{flalign} 

Given a set of aligned binary shape images, we learn the model of SBM for shape prior segmentation which maximize likelihood $p({\bf v};\theta)$ with respect to $\theta$. As proposed in \cite{c13}, the model parameters $\theta$ can be efficiently pre-trained at each layer greedily layer-by-layer. Joint training is then carried out to fine-tune the parameters and separate learning of local and global shape properties into the two hidden layers. 
 
\subsection{Incorporating the Shape Prior}
\label{sec:probrep}
If we were to optimize for only the first term of our objective function \ref{eq:nms_energy}, we would be solving an instance of a normal multi-region segmentation problem in which we partition the image into disjoint regions $R'_i$ associated with each object -- the possibly overlapping inferred boundaries $R_i$ would be playing no role. Since parametric representations of boundaries are typically insufficiently expressive and/or badly behaved for optimization purposes, in the literature, minimization of objective functions of this form is typically handled using the level set method \cite{c29}, where the interior of each region is represented as the positive locus of some continuous function of the image domain and a time parameter. For an overview of the level set method in the context of image processing, see \cite{c29}.

In the level set framework, region shape is commonly represented using a \emph{signed distance function}, that is, a function that gives the shortest distance to the nearest point on the boundary inside the region and its negative outside. The SBM used for the shape energy term, however, instead operates on a different continuous relaxation of region membership, namely a function $\q:\Omega\rightarrow [0,1]$ that assigns a probability that each pixel of the image is contained within the region in question. To be able to incorporate the shape energy term, we need to find a common representation, and reformulating the data term in terms of $\q$ is comparatively straightforward.


When working with such a function, a natural approach is to optimize on it directly and pick some cutoff probability if a binary notion of region membership is again required. This was first considered by Cremers, Schmidt and Barthel in \cite{c16} following \cite{c15} with the purpose of convexifying the space that optimization is performed over. 

Recall (\ref{eq:nms_energy}) that for a single region and its complement, the data term of the NMS functional took the following form:
\begin{eqnarray*}
E_{\mathrm{data}} &=& \int _{R} |{\bf u}-c_1|^2 dxdy + \int _{R^c} |{\bf u}-c_2|^2 dxdy
\end{eqnarray*}
Here, $c_1=c_1(R)$ and $c_2=c_2(R)$ are the average intensities in $R$ and $R^c$ respectively, and so in particular depend on $R$. 

Suggestively, we may reformulate this expression with a single integral taken over the whole image domain using the characteristic function $\chi_R$ of the region $R$ as follows: 
$$ E_{\mathrm{data}} = \int _{\Omega} |{\bf u}-c_1|^2 \chi_R + |{\bf u}-c_2|^2 (1-\chi_R) dxdy. $$

As $\bf q$ can be seen as a relaxation of $\chi_R$, this immediately suggests the following relaxed form in terms of ${\bf q}$:
\begin{eqnarray}
&E_{\mathrm{data}}({\bf q}) = \int _{\Omega} |{\bf u}-c_1|^2 {\bf q} + |{\bf u}-c_2|^2 (1-{\bf q})dxdy.& \label{eq:data-prob} 
\end{eqnarray}

\subsection{Decomposed Objective Function}
\label{sec:decomposed}
We will now show how to integrate the above shape representation with the shape prior model \ref{eq:spsd_energy2} to obtain an objective function for the problem that we may perform optimization on.

Noting how the data term for one region and the background in equation \ref{eq:data-prob} consist of a part associated with the region and one associated with its complement, we can decompose the data-based energy term to obtain a collection of coupled energy terms for each of the curve boundaries individually following the approach taken in \cite{c26}. We modify the NMS representation to keep track of the probability that each point is part of a given region, instead of the usual binary notion of region membership. Letting $\q_i:\Omega\to[0,1]$ be the membership probability function of the region $R_i$, we set 
\begin{flalign}
 &E_{data}(\q_i) & \label{eq:multi_func_rewritten} \\
 &= \int _{\Omega} \prod_{j=1}^{i-1} (1-\q_j) \left( |{\bf u}-c_i|^2 \q_i + (1-\q_i) \Phi_i \right) dxdy,& \nonumber \end{flalign}
where $\Phi_i$ is 
\begin{eqnarray}
\Phi_i &=& \sum_{j=i+1}^{n+1} \left( |{\bf u}-c_j|^2 \q_j \prod_{k=2}^{j-1} (1-\q_k) \right)
\end{eqnarray}
with $\q_{n+1}$ understood to be 1 everywhere; the product of $\q_i$s and $(1-\q_i)$s can be seen to be the natural generalisation of the characteristic function of the visible region $R'_i$ to the probabilistic relaxation.

We combine the shape prior term with this data term and reformulate our objective function (\ref{eq:nms_energy}) as follows:
\begin{flalign} 
&E_{SPSD} =  \sum_{i=1}^{N} \left(E_{data}({\bf q}_i)+\mu E_{shape}({\bf q}_i) \right) \nonumber\\
&=   \sum_{i=1}^{N} \int _{\Omega} |{\bf u}-c_i|^2 {\bf q}_i +  \Phi_i(1-{\bf q}_i) dxdy \nonumber\\
&-\mu ({\bf q}_i^{\mathrm{T}} W^1{\bf h}^1-{\bf q}_i^{\mathrm{T}}{\bf b}
  -\bf c^{1{\mathrm{T}}}h^1-h^{1\mathrm{T}}W^2 h^2-c^{2\mathrm{T}}h^2) \nonumber \\
&+\nu ||\nabla \q||_e.  
 \label{eq:spsd_energy2}
\end{flalign}
Note that we consider every region simultaneously while performing gradient descent. In particular, at any step, only the part of a given region that is taken to be unoccluded is compared to its mean luminance for the data agreement term; e.g. a hypothetical completely occluded region would then simply converge to the nearest plausible shape according to the shape prior with no regard for data agreement. 

\subsection{Minimizing the Energy Function}
When the learning model parameters are known, $E_{SPSD}$ has two kind of unknowns: the shapes ${\bf q_i}$ and the SBM related hidden units $\bf h^1$ and $\bf h^2$. Instead of addressing both together, we use an alternating minimization procedure. Each layer of hidden units can be computed by mean-field approximate inference, just as done for DBM \cite{c15}. Observe that energy functional \ref{eq:spsd_energy2} with respect to each shape ${\bf q_i}$ is a convex functional, and hence can be solved using the Split Bregman method to obtain a global minimizer. 
We apply the Split Bregman method \cite{c19} to solve the minimization problem of step 5 in Algorithm \ref{<Multiple Object segmantation with depth>}, adapting the implementation in \cite{c20} with respect to $\q$, and perform exhaustive search to find the minimizing translation and scaling of $\mathbf{W}$.

\begin{algorithm}
\caption{Multi-region segmentation with depth}
\label{<Multiple Object segmantation with depth>}
\KwIn{an image $u$, model parameters $\theta = \{\bf \tilde{W}^1, W^2, c^1, c^2, b\}$, and a test image $\bf u$}
\nl Initialize {$\{\q_i^0\}_{i=1}^{N-1}$}\\
\Repeat{\nl$\sum_{i=1}^{N-1} \| {\q_i}^{k}-{\q_i}^{k-1}\|<\epsilon$}{
\nl$k\leftarrow k+1$\\
\For{$i=1\ldots N-1$}{
\nl${\bf h^1}\leftarrow \sigma({(\q_i^{k-1})^{\mathrm{T}}\tilde{W}^1+W^2 h^2+c^1})$\\
\nl${\bf h^2}\leftarrow \sigma({\bf h^{1\mathrm{T}}W^2+c^2})$\\
\nl${\q_i^{k}}\leftarrow \arg\min_{\q,\mathbf{W}}( E_{data}(\q)-\q^{\mathrm{T}}\tilde{\mathbf{W}}^1 h^1-{\q}^{\mathrm{T}}b )$\\
}
}
\nl\KwRet{$\{\bf q_i\}_{i=1}^{N-1}$}
\end{algorithm}

The initialization step in Algorithm 1 is performed by running one iteration of the Split Bregman minimization without a shape prior, i.e. setting $E_{SBM}({\bf v,h^1,h^2}; \theta)=0$. We have a $O(n^2)$ dependency on number of regions, as the shape prior is applied independently, and each $E_{data}(\mathbf{q}_i)$ in equation (9) takes $O(n)$ to calculate naively. This may easily be improved to $O(n)$ by carrying through partial products.
The dependency on image size is also $O(n^2)$, as both the Split Bregman iteration and our updates of $\mathbf{q}$ are local.


\section{Experiments}
 All experiments were run in MATLAB on a PC with a 2.30 GHz Intel Core i7-3610QM Processor and 8GB of RAM. To implement the Split Bregman $\arg\min$ in Algorithm 1, we used the MATLAB native code wrapper to interface with a modified version of the code provided by T. Goldstein as a supplement to \cite{c20}.

\begin{figure}[t]
\begin{center}
\includegraphics[width=1.0\linewidth]{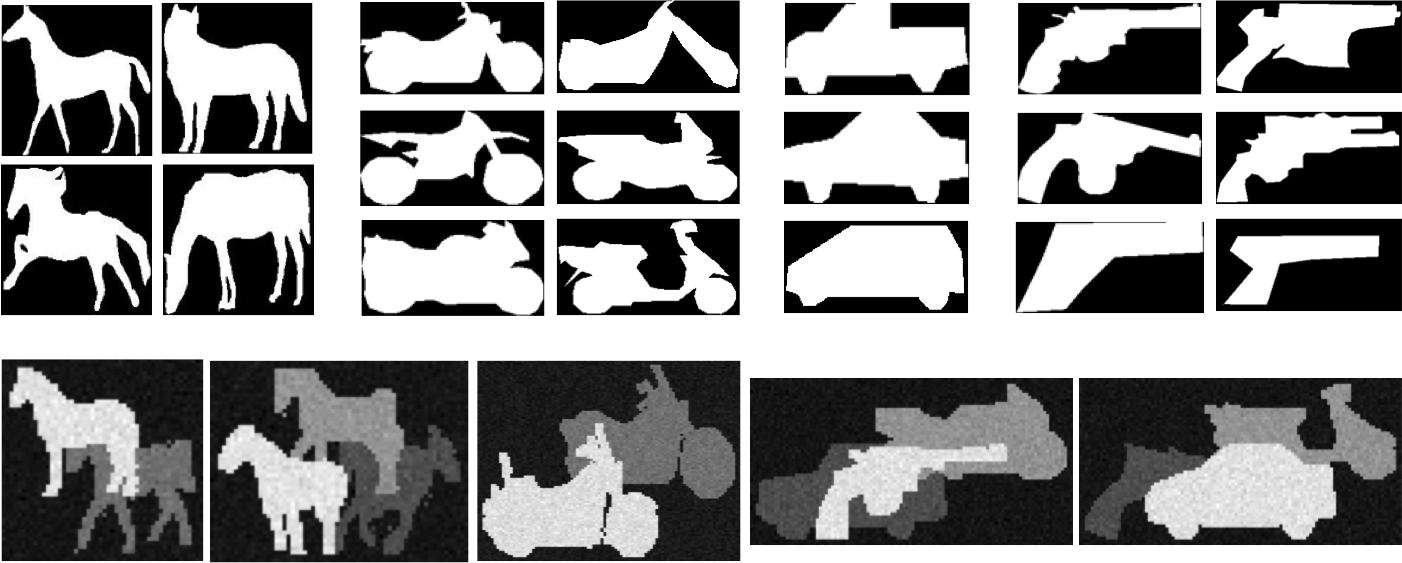}
\end{center}
\vspace{-0.5em}
\caption{Some examples of shape dataset (top) and the synthetic images (bottom). The synthetic image is generated by locating two or three resized binary shapes and adding noise. The combinations of the shapes used in the synthetic images are randomly selected. }
\label{fig:synthetic_samples}
\end{figure}

We evaluate the effectiveness of our algorithm on several datasets which consist of binary segmentation masks representing an object silhouette (Fig. \ref{fig:synthetic_samples}(top)). Through the evaluation, those binary images are cropped and normalized to the specific size, and about half of the images are used for training, and the rest of them for testing. In the SBM training phase, pre-training requires 3000 and 1000 epochs for the first and second layers. In addition, global training is performed for 1000 epochs. 

As corresponding algorithms using Split Bregman optimization with Markov Random Fields (MRF), Factor Analysis \cite{c34}, and the Restricted Boltzmann Machine (RBM) have been investigated in \cite{c14}, we restrict ourselves to evaluating our algorithm's performance on each input in comparison to that of the single-region shape prior segmentation algorithm of Chen et al. (CVPR 2013) \cite{c15}, with the DBM model it employs replaced with our SBM model to ensure that only the effect of treating occlusion differently is captured. We employ this single-region algorithm for multi-region segmentation in a natural way by considering every object separately in turn and treating the rest of the image as the ``background''.  In the tables, this algorithm is referred to as ``SP (single)''. We also consider the results obtained by an analogous algorithm that does not employ the machine-learned shape priors at all (referred to as ``no SP'') for perspective. 
 
We evaluate the algorithms on both synthetic images and real images. The synthetic images are generated by overlapping two or three binary shapes selected uniformly at random from the test set. We also added Gaussian noise to each image. Fig. \ref{fig:synthetic_samples} shows some examples of the synthetic images used in the experiment. 

For the synthetic images, the accuracy of the segmentation can be evaluated by comparing it to the ground truth segmentation, i.e. the individual silhouettes that were composed to form the test image. Since each segmented region corresponding to ${\bf q_i}$ has a pixel value in $[0,1]$, where each value represents the probability that the pixel is in the region, we simply take as the region those pixels having value larger than 0.5 to compare with the binary segmentation mask. We use two common metrics to assess segmentation quality: the average pixel accuracy (AP), of foreground and background classification, and the foreground intersection-over-union score (IoU), defined as 
$O_{iou}(S_{p}, S_{q}) = \frac{S_{p} \cap S_{p}}{S_{p} \cup S_{p}}$
where $S_{p}$ and $S_{q}$ are the sets of ground truth and predicted foreground pixels.

\subsection{Single-Class Dataset}
We evaluate the basic performances on two datasets. 

\smallskip\noindent\textbf{Caltech-101 motorbikes}: The first dataset we investigated was 798 motorbike silhouettes from Caltech-101 dataset \cite{c24}, capturing motorbike from side. We used 399 images for training, and the rest of them for testing. The images are cropped and normalized to 64 $\times$ 64 pixels. We trained an SBM with $d=4$, and 1200 and 50 units for $h^1$ and $h^2$ respectively. 

\smallskip\noindent\textbf{Weizmann Horses}: The second dataset is the Weizmann horse dataset \cite{c25} consisting of 328 images, all of horses facing left, but with a high variety of poses. We created a training set of 164 images from this dataset, cropped and normalized to 32 $\times$ 32 pixels. We trained an SBM with $d=4$, and 2000 and 100 units for $h^1$ and $h^2$ respectively. 

\begin{figure}[t]
\begin{center}
\includegraphics[width=0.95\linewidth]{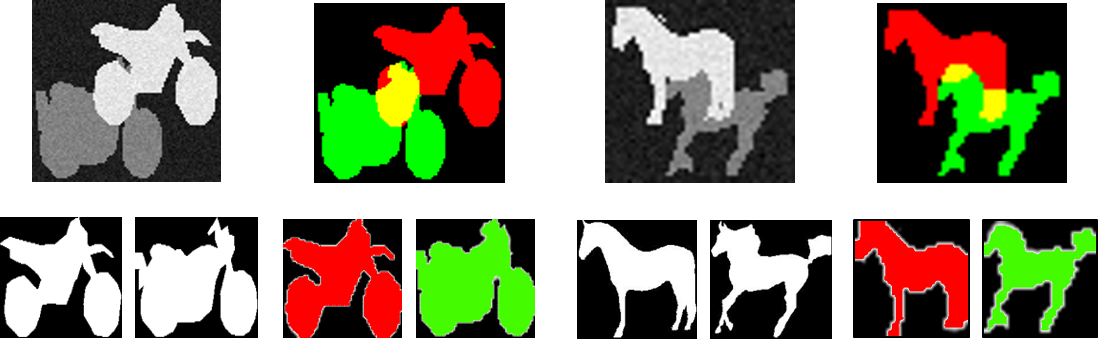}
\end{center}
\vspace{-4mm}
\hspace{6mm}
(a)\hspace{16mm}
(b)\hspace{18mm}
(c)\hspace{16mm}
(d)
\begin{center}
\includegraphics[width=1.0\linewidth]{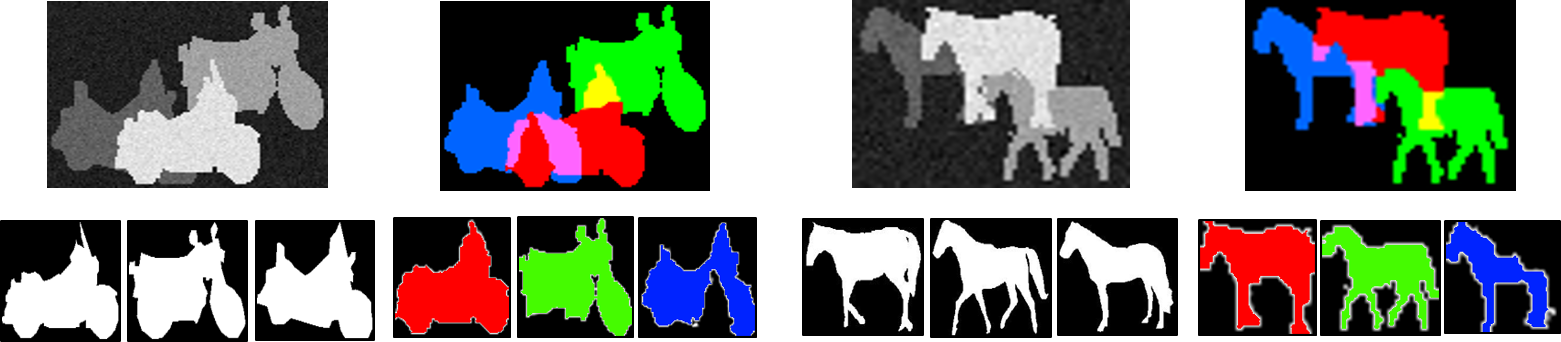}
\end{center}
\vspace{-3mm}
\hspace{6mm}
(e)\hspace{16mm}
(f)\hspace{18mm}
(g)\hspace{16mm}
(h)
\vspace{1mm}
\caption{Segmentation results with a single object class for two (a-d) and three (e-h) regions. Some input images, together with the respective ground truth shapes, are shown in (a)(c) and (e)(g). The composite segmentation outputs (with the regions being rendered, from front to back, in red, green and blue), along with the recovered shapes, can be seen in (b)(d) and (f)(h).}
\label{fig:result_basic}
\end{figure}

Example images of results for motorbikes and Weizmann horses are shown in Fig \ref{fig:result_basic}.
The shape prior based algorithms can be seen to benefit from the power of the SBM model: even if one of the main components of the object, such as a handle of the motorbike or a leg of the horse, is missing, the proposed method still determine an proper boundary of the object recovering its plausible shape. Red indicates the pixels within the first object from the front, green indicates the second object from the front, and blue indicates the third object from the front.  

\begin{figure}[t]
\begin{center}
\includegraphics[width=0.80\linewidth]{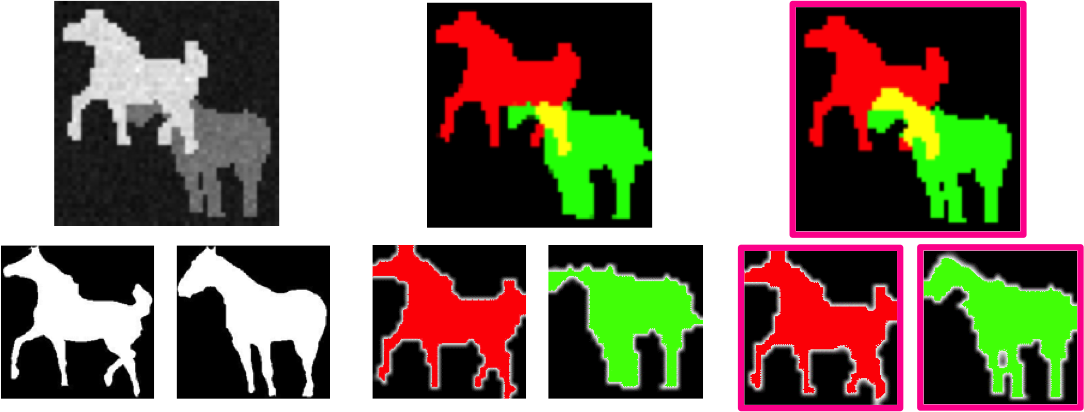}
\end{center}
\vspace{-4mm}
\hspace{10mm}
(a)\hspace{25mm}
(b)\hspace{23mm}
(c)
\begin{center}
\includegraphics[width=0.80\linewidth]{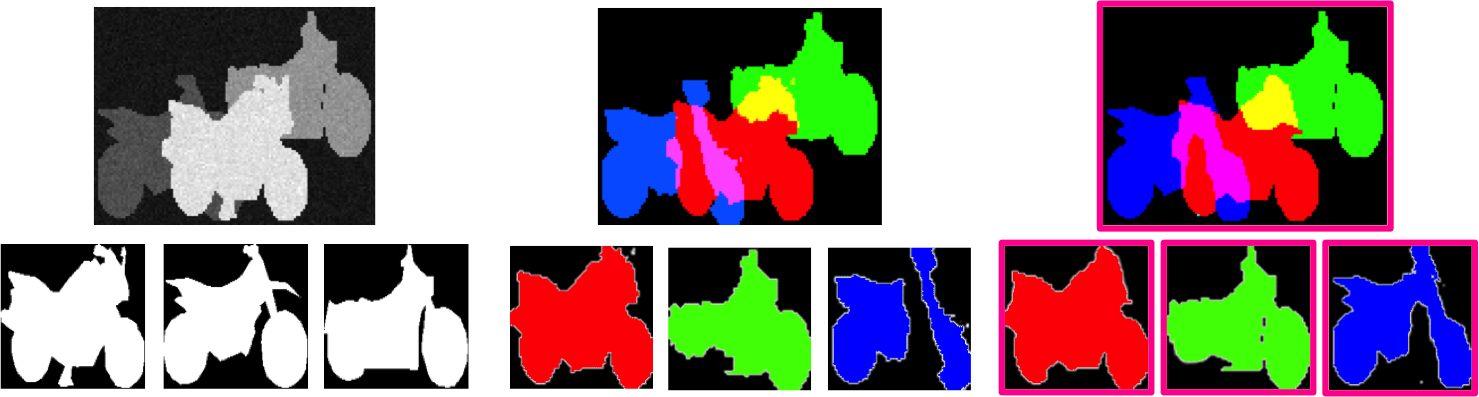}
\end{center}
\vspace{-3mm}
\hspace{10mm}
(d)\hspace{25mm}
(e)\hspace{23mm}
(f)
\vspace{1mm}
\caption{Segmentation comparison of our method (multi-region segmentation) vs. single-region segmentation running several times: Original synthetic images are shown in (a),(d). (b)(e) and (c)(f) show the segmentation results and the corresponding shapes of single-region segmentation, our method, respectively. Our results are encircled by red line. }
\label{fig:result_multi_vs_single}
\end{figure}

The qualitative and quantitative comparison of our algorithm with ``SP (single)'' and ``no SP'' can be seen in Fig. \ref{fig:result_multi_vs_single} and Table \ref{table:result_singleclass}. Testing was performed with 399 motorbike images and 164 horse images.

We expected the single-region approach to suffer in occluded regions as $E_{data}$ is raised by the mismatch in pixel intensity.
The measurements appear to confirm this prediction -- our method exhibits significantly improved scores for background regions across the board, and visual inspection confirms that this is largely due to improved accuracy in occluded parts. 
The keen observer may note that using a shape prior results in slight loss of accuracy for the front-most region. This is unsurprising: without the shape energy term, the contour is free to evolve to exactly match the uniformly colored region, but artifacts of the SBM shape model may lead to slight deviations having lower total energy.

\begin{table*}
\begin{center}
  \begin{tabular}{| c | c |  c | c | c || c |  c | c | c | c | c |} \hline
      \multicolumn{1}{| c |}{method} &  \multicolumn{2}{| c |}{AP}
      & \multicolumn{2}{c ||}{IoU} & \multicolumn{3}{| c |}{AP}
      & \multicolumn{3}{| c |}{IoU} \\ \cline{2-11} 
         &  \multicolumn{1}{| c |}{region1} &  \multicolumn{1}{| c |}{region2}
      &  \multicolumn{1}{| c |}{region1} & \multicolumn{1}{c ||}{region2} &  \multicolumn{1}{| c |}{region1} &  \multicolumn{1}{| c |}{region2} &  \multicolumn{1}{| c |}{region3} & \multicolumn{1}{| c |}{region1}  & \multicolumn{1}{| c |}{region2} & \multicolumn{1}{| c |}{region3} \\ \hline
   no SP         & {\bf 100}&88.67&{\bf 100}&78.76      & {\bf 100}&92.77&71.48&{\bf 99.99}&86.23&49.51 \\   
      SP (single)& 98.94 & 89.35&98.27&80.95            & 98.79&92.88&82.87&98.08&87.24&71.39 \\ 
      SP (multi) & 98.86&{\bf 94.04}&98.13&{\bf 89.52}  & 97.88&{\bf 96.25}&{\bf 86.98}&96.64&{\bf 93.24}&{\bf 78.41} \\ \hline
      no SP      & {\bf 99.92}&80.95&{\bf 99.83}&65.70  & {\bf 99.97}&81.98&67.46&{\bf 99.94}&69.27&41.94 \\ 
      SP (single)& 98.69 & 82.58&97.61&69.12            & 98.29     & 82.70&69.97&96.88&70.42 &47.91  \\
      SP (multi) & 98.36 &{\bf 87.62}&97.01&{\bf 77.96} & 97.65&{\bf 86.00}&{\bf 71.39}&95.71&{\bf 75.89}&{\bf 50.36} \\
      \hline      
  \end{tabular}
  \end{center}
  \caption{Results on the single class dataset: bike(top), horse(bottom). All standard deviations are bounded above by $\pm 0.23$.}
  \label{table:result_singleclass}
\end{table*}

\begin{table*}
\begin{center}
  \begin{tabular}{| c | c |  c | c | c || c |  c | c | c | c | c |} \hline
      \multicolumn{1}{| c |}{method} &  \multicolumn{2}{| c |}{AP}
      & \multicolumn{2}{c ||}{IoU} & \multicolumn{3}{| c |}{AP}
      & \multicolumn{3}{| c |}{IoU} \\ \cline{2-11} 
         &  \multicolumn{1}{| c |}{region1} &  \multicolumn{1}{| c |}{region2}
      &  \multicolumn{1}{| c |}{region1} & \multicolumn{1}{c ||}{region2} &  \multicolumn{1}{| c |}{region1} &  \multicolumn{1}{| c |}{region2} &  \multicolumn{1}{| c |}{region3} & \multicolumn{1}{| c |}{region1}  & \multicolumn{1}{| c |}{region2} & \multicolumn{1}{| c |}{region3} \\ \hline
      no SP & {\bf 100}&84.35&{\bf 100}&71.89 & {\bf 99.92}&85.30&63.10&{\bf 99.91}&73.80&32.53 \\
SP (single) & 98.56 & 85.88&97.93&76.74& 98.43&85.59&70.58&97.59&76.96&57.96 \\
SP (multi) & 98.22&{\bf 86.39}&98.22&{\bf 86.39} & 98.03&{\bf 92.04}&{\bf 77.96}&97.00&{\bf 86.76}&{\bf 66.35}  \\\hline    
  \end{tabular}
  \end{center}
  \caption{Results on the multiclass dataset with three object categories from Caltech-101. All std. deviations are bounded above by $\pm 0.16$.}
  \label{table:result_multiclass}
\end{table*}

\begin{figure}[h]
\begin{center}
\includegraphics[width=0.95\linewidth]{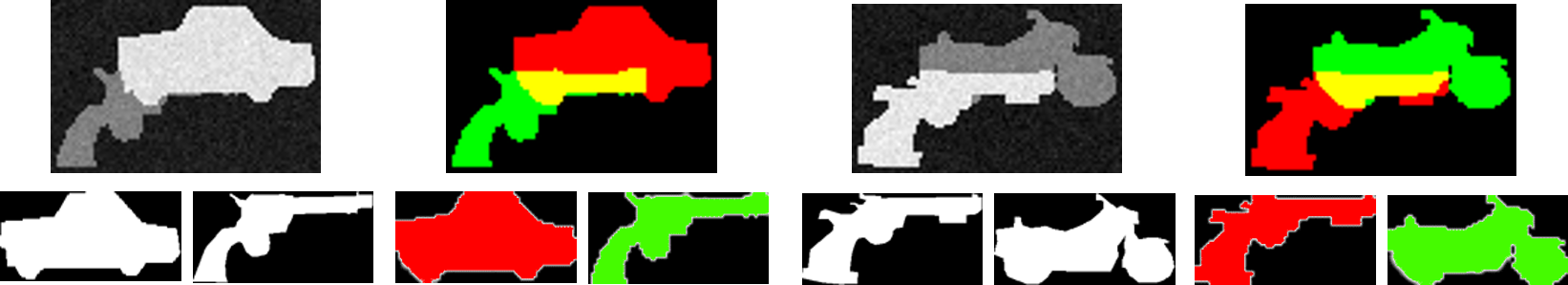}
\end{center}
\vspace{-4mm}
\hspace{9mm}
(a)\hspace{15mm}
(b)\hspace{16mm}
(c)\hspace{15mm}
(d)
\begin{center}
\includegraphics[width=0.75\linewidth]{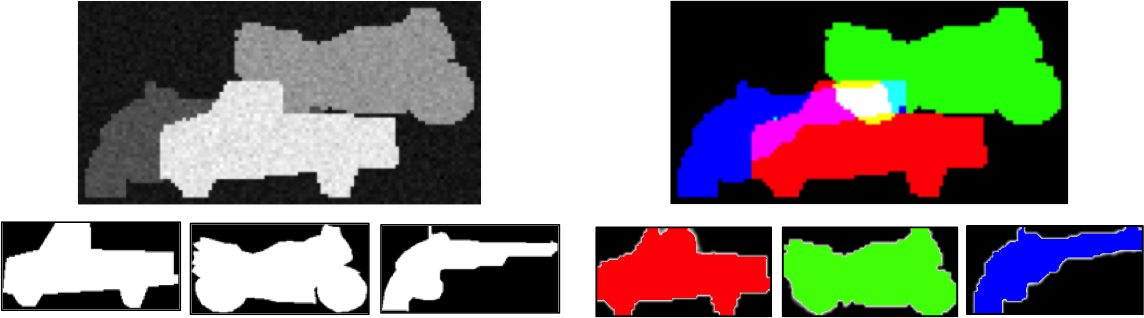}
\end{center}
\vspace{-1.0em}
\hspace{20mm}
(e)\hspace{36mm}
(f)
\begin{center}
\includegraphics[width=0.75\linewidth]{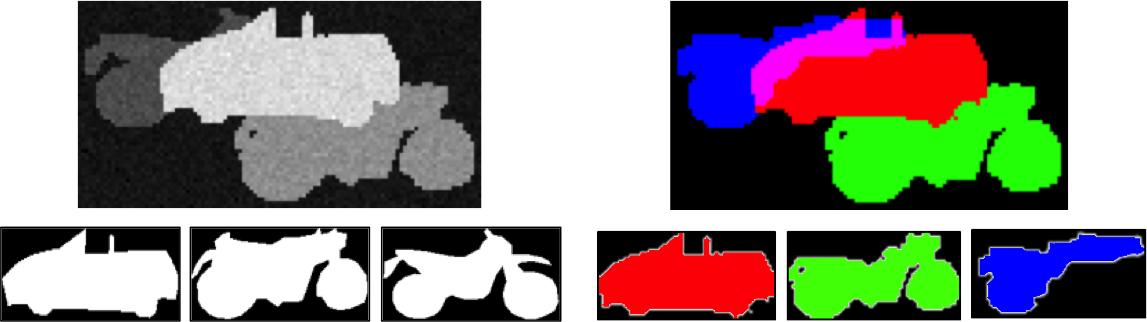}
\end{center}
\vspace{-1.0em}
\hspace{20mm}
(g)\hspace{36mm}
(h)
\caption{Segmentation results for multiple object classes. Input images and ground truth segmentations shown in (a),(c), (e) and (g). Segmentation results are in (b),(d), (f) and (h). (h) depicts an instance of segmentation failing: the third object, a partially occluded motorbike, is recovered as a revolver. }
\label{fig:result_multiclass}
\end{figure}

\begin{figure}
\begin{center}
\scalebox{0.8}{
  \begin{tabular}{| c | c | c | c | c  |} \hline
    \multicolumn{1}{| c |}{dataset} &  \multicolumn{2}{| c |}{single}
      & \multicolumn{2}{| c |}{multi} \\ \cline{2-5} 
        &  \multicolumn{1}{| c |}{2 regions} &  \multicolumn{1}{| c |}{3 regions}  & \multicolumn{1}{| c |}{2 regions} & \multicolumn{1}{| c |}{3 regions} \\ \hline
   bike  &2.39($\pm 0.80$)&5.43($\pm 1.00$)&3.50($\pm 3.04$)&7.63($\pm 3.32$) \\ \hline
   horse  &1.74($\pm 0.62$)&4.03($\pm 1.74$)&1.05($\pm 0.73$)&4.36($\pm 2.08$) \\  \hline
   Caltech  & 2.28($\pm 0.81$)&5.31($\pm 0.44$)&1.45($\pm 1.29$)&8.32($\pm 3.37$) \\ \hline
  \end{tabular} }
  \end{center}
\caption{Average runtime per instance for the three synthetic datasets, measured on our test system, in seconds.}
\end{figure}

\subsection{Multi-Class dataset}
Next, we shows the results of multi region segmentation with multiclass shape prior.  We tested how well our method infers highly occluded object shape without any information about the category of each object, leaving it to the shape model to discover the most likely object categories from input image.  
For this purpose, we trained SBM on a combination of the images of motorbike, revolver, and car of side view from Caltech-101 dataset \cite{c24}. The training dataset contains 60 cars in profile, 40 revolvers, and 300 motor bikes for a total 400 images. The images are cropped and normalized to 64$\times$32 pixels. A SBM with $d=4$, and 1500 and 300 units for $h_1$ and $h_2$ was jointly trained without information about image class.  

Example segmentation results can be seen in Fig. \ref{fig:result_multiclass}; quantitative results from 599 synthetic images can be found in Table \ref{table:result_multiclass}. We observe that our algorithm largely identifies the right object category and converges to the appropriate shape.
There are, however, some cases where the occlusion of the shape is too heavy to determine the plausible object category. In the false example in Fig. \ref{fig:result_multiclass}, a partially occluded motorbike is recovered as a revolver. 

\begin{figure*}[t]
\begin{center}
\includegraphics[width=0.90\linewidth]{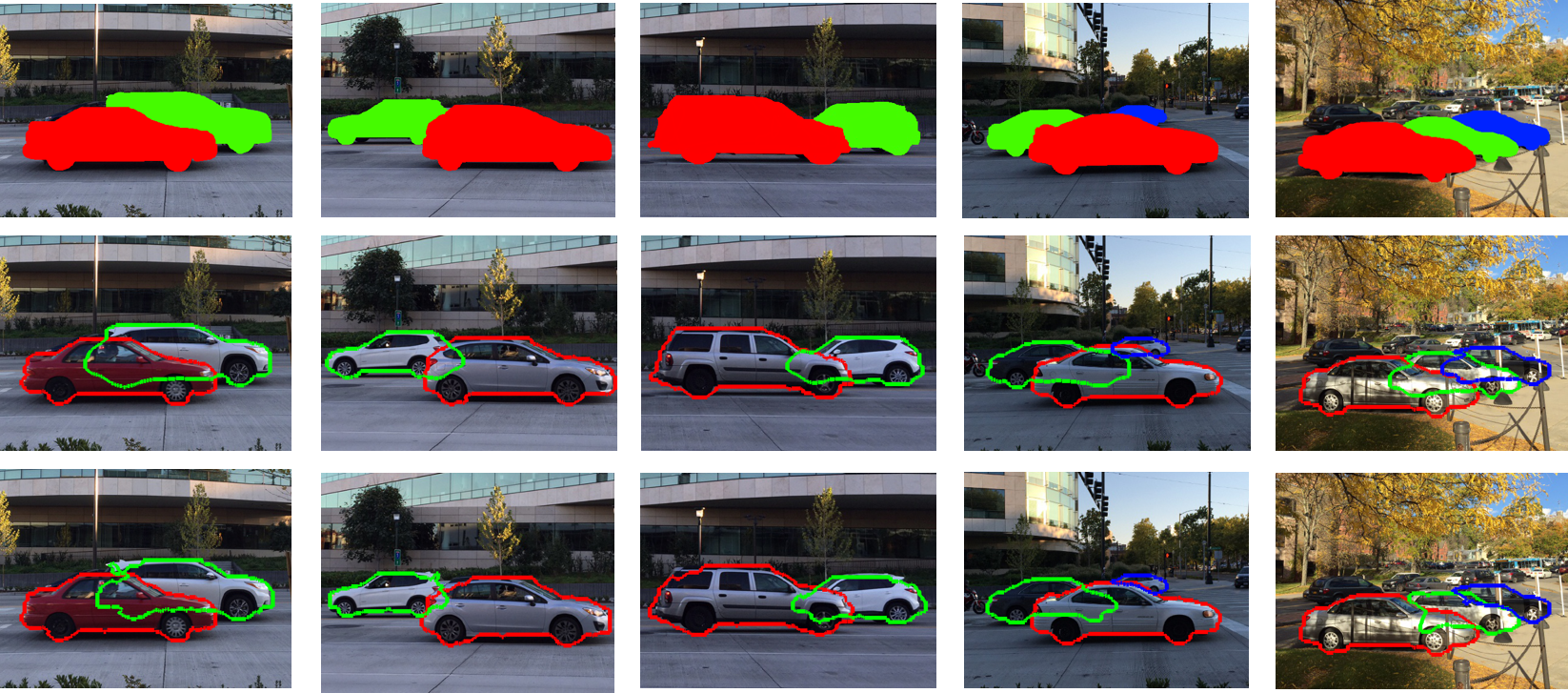}
\end{center}
\caption{Qualitative test results on enhanced real-world images. Top row: Input. Middle row: Results of our algorithm, overlaid onto the original image. Bottom row: Results of the single-region shape prior algorithm.}
 \label{figure:result_realimage}
\end{figure*}

\begin{figure*}
\begin{center}
\includegraphics[width=0.90\linewidth]{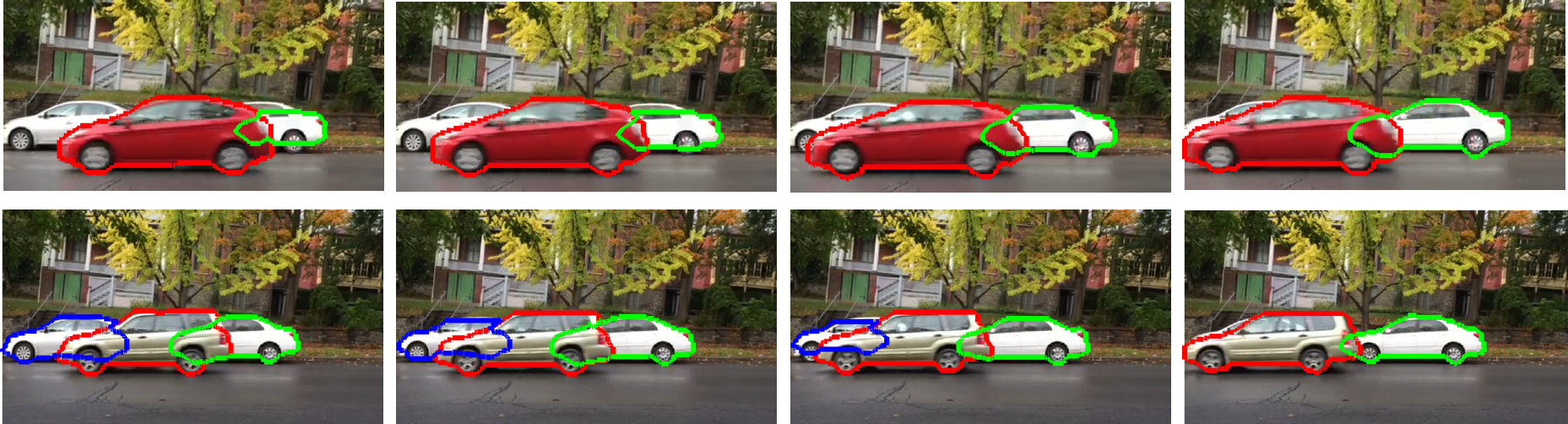}
\end{center}
\caption{Qualitative test results on a few subsequent frames of enhanced real-world images of a moving car.}
 \label{figure:result_realimageseq}
\end{figure*}

\subsection{Enhanced Real-World Images}
Next, we tested the proposed method on real images. 
In application, the target object often has similar shapes in different scales. The data set we tested here contains partially occluded cars with different direction, scale, and location. 
We used 60 side car shapes (out of a total 123) from the Caltech 101 dataset, with all cars facing left but not necessarily being of the same type.
 In order to segment a car facing the other direction simultaneously, we construct another 60 training shapes by flipping the previous training shapes horizontally.
We cropped and normalised all shapes to $64\times 32$ pixels, and trained an SBM with 1200 units in the first and 50 units in the second hidden layer on this extended training set with a total 120 shapes.
The initial segmentation of visible part of the object is manually given.
Since the aspect ratio of the car shapes is changing in the training phase, we normalized input images so that the size of the car located forward will be ratio of 64:32 in segmentation process.
Here, too, we find that our algorithm outperforms the reference in the same way (Fig. \ref{figure:result_realimage}).

We have found that the performance depends on the occlusion ratio. If more than half of the shape is occluded, our algorithm often mistakes the the visible part for one belonging to a smaller copy of the object. Some consecutive frames of a video of a moving car have been segmented in Fig. \ref{figure:result_realimageseq} to show how segmentation performance declines as more and more of an object is occluded.

One possible approach to overcome this limitation is to integrate {\it context}, such as scale (objects further away should be smaller), orientation (should be similar each other among existing objects), or physical plausibility, in our algorithm.

\section{Conclusion}

We have presented a variational algorithm for multi-region segmentation of partially overlapping objects. Our algorithm uses prior information about object shapes modeled by a Shape Boltzmann Machine (SBM), and integrates it with a data-driven term. The data-driven term is made to depend only on image pixels assumed to be relevant, based on a probabilistic representation of region membership compatible with the SBM. On one hand, this enables it to deal with noise and damage in visible parts in the usual way; on the other hand, it also freely infers the shape of parts that are missing in the image due to being occluded, without having to compete with irrelevant data originating from occluding objects. The latter enables it to exhibit significant improvements over a variational algorithm representative of the state of the art, which treats occluding objects as damage. 

We believe that our data supports the thesis that accounting for occluding objects is crucial to approaching human performance at multi-region segmentation. Since many subpar segmentations produced by our algorithm exhibit shapes that get good scores from the SBM but do not actually depict plausible silhouettes of the object in question, performance may likely be improved further by employing a more advanced model for the shape prior.


\section*{Acknowledgements}
The authors were partially supported by the following grants: NSF CCF-1214844, MURI FA9550-12-1-0040 and ARO W911NF-09-1-0281. We would like to thank the Cornell AMP Lab, Kevin Matzen, Kyle Wilson, the Cornell Graphics and Vision group and the anonymous reviewers for support and helpful comments.

{\small
\bibliographystyle{ieee}
\bibliography{egbib}

}

\end{document}